\documentclass[a4paper, UKenglish, cleveref, autoref, thm-restate, authorcolumns]{lipics-v2021-poster}

\title{\textbf{\textit{In-Context Learning to Assess Built Environment Impacts on Perceived Neighborhood Walkability Among Mobility-impaired Older Adults}}} 

\titlerunning{Decoding Perceived Walkability via In-Context Learning} 

\author{Houhao Liang}{CNRS@CREATE, 1 Create Way, Singapore}{houhao.liang@cnrsatcreate.sg}{0000-0003-3491-3281}{}

\author{Kresimir Friganovic}{Future Health Technologies, Singapore-ETH Centre, 1 Create Way, Singapore}{kresimir.friganovic@sec.ethz.ch}{}{}

\author{Joanne Kua}{Institute of Geriatrics and Active Ageing, Tan Tock Seng Hospital, Singapore}{joanne_kua@ttsh.com.sg}{}{}

\author{Noor Hafizah Ismail}{Institute of Geriatrics and Active Ageing, Tan Tock Seng Hospital, Singapore}{noor_hafizah@ttsh.com.sg}{}{}

\author{Su Su}{Geriatric Medicine, Khoo Teck Puat Hospital, Singapore}{su.su@ktph.com.sg}{}{}

\author{Bryan Yijia Tan}{Department of Orthopedic Surgery, Woodlands Health Campus, Singapore}{bryan.tan@nhghealth.com.sg}{}{}

\author{Navrag B. Singh}{Future Health Technologies, Singapore-ETH Centre, 1 Create Way, Singapore}{navrag.singh@sec.ethz.ch}{https://orcid.org/0000-0001-8074-041X}{}

\author{Panos Mavros\footnote{Corresponding author}}{i3 (UMR-9217 CNRS), Department of Social and Economic Sciences, Télécom Paris, Institut Polytechnique de Paris, Palaiseau, France}{panos.mavros@telecom-paris.fr }{[https://orcid.org/0000-0002-3027-3072]}{}

\authorrunning{H. Liang et al.} 

\Copyright{Houhao Liang} 



\category{Poster} 

\relatedversion{} 

\funding{This research was supported by the Singapore National Research Foundation under Intra-CREATE Thematic Grant (NRF2022-THE004-004).}

\acknowledgements{The authors would like to thank Tan Chun Yue for assistance with study conduct and research administration. Special thanks are also due to the following individuals for their valuable contributions to the experimental setup and logistical coordination of data collection: Gillian Chew Long Szu, Tan Jia Yi, and Elaine Tan Jie Shuang from Woodlands Health; Ken Lim Sze Yang and Tong Pei Ling from Tan Tock Seng Hospital; and Cherlyn Lee and Haw Hui Yi from Khoo Teck Puat Hospital.} 

\EventEditors{Sabine Timpf, Gabriele Filomena, Armand Kapaj, Rui Zhu, Nicholas Giudice, and Ed Manley}
\EventNoEds{6}
\EventLongTitle{17th International Conference on Spatial Information Theory (COSIT 2026)}
\EventShortTitle{COSIT 2026}
\EventAcronym{COSIT}
\EventYear{2026}
\EventDate{September 22--25, 2026}
\EventLocation{York, UK}
\EventLogo{}
\SeriesVolume{}
\ArticleNo{}
\DOI{}

\begin{document}

\maketitle
\begin{abstract}
As global populations age, enhancing neighborhood walkability through inclusive urban design is important to mitigate the built environment (BE) barriers that discourage physical activity and social participation among older adults. This study investigates the utility of In-Context Learning (ICL), leveraging the transformer-based foundation model TabPFN to decode how BE features function as determinants of perceived walkability, as measured by the Neighborhood Environment Walkability Scale (NEWS-A) survey. Using a small-scale dataset (N=257) of a unique demographic, older adults with knee osteoarthritis or a history of falls, TabPFN achieved a Macro F1 score of 54.89\% on walkability perceptions discretized into Low, Neutral, and High levels via equal binning. This result  outperformed optimized, grid-searched baselines, including Random Forest (45.85\%) and XGBoost (50.56\%). To interpret these results, we employed Shapley Interaction Quantification (SHAP-IQ) to identify the hierarchical importance of feature synergies. The preliminary results revealed that the model's predictive logic is primarily driven by higher-order interactions. For example, the interplay between average street circuity and the ratio of drivable roads emerged as the primary discriminator of walkability perception, and neighborhood greenery was found to hold substantial predictive weight only when coupled with an individual’s fear of falling or perception of age-friendliness. Overall, ICL via TabPFN demonstrates superior performance on small-scale datasets, enhancing the fidelity of the resulting interpretive insights. Furthermore, SHAP-IQ provides a synergistic perspective on how higher-order feature interactions drive the model's predictions.

\end{abstract}
\section{Introduction}
As global populations age, the intersection of the built environment (BE) and geriatric health has become a critical focal point for urban policy \cite{who2021decade}. For  mobility-impaired older adults, particularly those managing conditions such as knee osteoarthritis (OA) or a history of falls, the urban fabric is far from a neutral backdrop. Instead, it represents a dynamic landscape of barriers and supports where urban configurations can either facilitate mobility or act as considerable physical constraints \cite{TORKU2022109533}. Therefore, these BE features dictate the individual’s capacity for movement, directly influencing their willingness to engage in physical activity and social participation.

Existing studies analyzing the impacts of the BE on older adults primarily rely on linear regression \cite{CARSON2023103036, JIANG2025105587}. However, these models are fundamentally constrained by assumptions of linearity and often fail to capture the complex interaction effects inherent in human-environment dynamics. In reality, BE features often interact in non-linear and synergistic ways, resulting in a compounding effect on perceived walkability. To uncover these complexities, recent studies in walkability analysis have increasingly leveraged inductive Machine Learning (ML) models such as Random Forest (RF) and XGBoost \cite{YANG2024102087}, with SHAP for results interpretation. However, the performance and subsequent fidelity of these models are often contingent upon the availability of large-scale datasets. Because their underlying logic involves a stepwise partitioning of features, the models are most robust when supported by the extensive sample sizes that allow for stable decision boundaries without risking overfit.

To address this issue, In-Context Learning (ICL), a learning paradigm originally popularized by Large Language Models, has emerged to perform approximate inference via transformer-based architectures \cite{hollmann2022tabpfn}. In contrast to traditional inductive ML models that require iterative gradient updates to learn patterns from scratch, ICL operates by taking a small dataset as a Context Set. The frozen weights of a foundation model pre-trained on large tabular data function as a generalized inference engine, providing the intrinsic rules for evaluating feature distributions and correlations. During inference, the self-attention mechanism can dynamically attend to the provided context, generating predictions for new instances in a single forward pass without any further parameter optimization. Hence, this learning paradigm mitigates the constraints of the small-data bottleneck while minimizing the reliance on extensive hyperparameter tuning.

Given that the BE represents a complex environment that is cognitively perceived and decoded by the individual, this research serves as an exploratory investigation into the utility of ICL for decoding its impact on on a uniquely vulnerable demographic: older adults with knee OA or a history of falls. Their subjective perception of neighborhood walkability is operationalized through the Neighborhood Environment Walkability Scale (NEWS-A) survey. To analyze this, this study employs TabPFN (Tabular Prior-Data Fitted Network), a transformer-based foundation model that implements ICL and is pre-trained on millions of synthetic tabular datasets \cite{hollmann2022tabpfn}, to perform transductive inference. Finally, Shapley Interaction Quantification (SHAP-IQ) \cite{muschalik2024shapiq} is applied to to interpret the results and identify the synergistic interactions.

\section{Methodology}
\subsection{Study Population}
The study cohort initially comprised 373 mobility-impaired older adults recruited in collaboration with Woodlands Health, Tan Tock Seng Hospital, and Khoo Teck Puat Hospital (Singapore). Individual clinical profiles were recorded, specifically focusing on physical vulnerabilities: Knee OA, history of falls within in two years, or a comorbidity of both.  Following data cleaning and the exclusion of incomplete responses, a final valid dataset of 257 participants was utilized for this study (mean age = 70.76 years; 68.87\% female). The cohort included individuals with knee OA (40.86\%) and a history of falls (38.52\%), with 20.62\% having both conditions. This study has been approved by the local institutional review board (IRB-2024-818). 

\subsection{Study design}
Participants completed a comprehensive set of psychometric and mobility surveys, including the Life-Space Assessment (LSA), Falls Efficacy Scale (FES), Age-Friendly Environment Assessment Tool (AFEAT), NEWS-A, and Patient Health Questionnaire-9 (PHQ-9). For the purposes of this study, the NEWS-A sum score serves as the primary target variable, representing the subjective perception of neighborhood walkability. 

To enable spatial analysis, participants' residential address (i.e. postcode) were geocoded using Singapore’s unique six-digit postal codes. This allowed to calculate multiple BE metrics relative to each participant's exact location within each neighbourhood. 

\subsection{Explanatory variables}
The BE feature selection is theoretically grounded in the 5D’s walkability framework (Design, Density, Diversity, Destination, and Distance to transit) \cite{ewing2010travel}. Given the hyper-density of the Singaporean urban fabric and established literature on senior mobility, we utilized a 400m radius buffer around each residential point to calculate morphological features, representing the immediate walkable neighborhood \cite{cerin2020urban}. For accessibility metrics, we moved beyond simple Euclidean proximity. We calculated network distances and circuity that is the ratio of network distance to Euclidean distance, to capture the actual navigational effort required by pedestrians. Routing was performed via the OneMap API (the official geospatial platform of the Singapore Land Authority), which provides higher accuracy for local navigation, including the consideration of pedestrian-only paths and overhead bridges.

A key challenge in high-density contexts is that evaluating accessibility based on individual, isolated Point of Interest (POI) can be misleading. Given the Singapore's land-use zoning framework, urban services are heavily integrated into centralized community hubs and designated blocks. Consequently, pedestrian trips are typically oriented toward these aggregate functional zones rather than standalone POI coordinates. Hence, we utilized land-use blocks (e.g., commercial, institutional, and recreational) as the primary proxies for destinations. 

Furthermore, we incorporate individual-level clinical context alongside objective BE features to create a holistic profile of the study participants. By integrating clinical indicators of mental and physical health (e.g., PHQ-9 and LSA) with high-resolution environmental data, the resulting feature set enables the model’s attention mechanism to attend to the complex interactions between an individual's physiological frailty and the specific morphological barriers within their neighborhood BE. Following a multicollinearity filtering process to ensure statistical robustness, the final feature set is organized into the categories presented in Table \ref{tab:features}.

\begin{table}[htb]
\centering
\caption{Integrated Feature Set (independent variables) for Decoding Perceived Neighborhood Walkability}
\label{tab:features}
\begin{tabular}{lp{0.3\linewidth}p{0.3\linewidth}}
\hline
\textbf{Category} & \textbf{Features} & \textbf{Description / Source} \\ \hline
\textbf{Morphologic} & Intersection Density, Average Circuity, Greenery Ratio, Population Density, Covered Walkway Ratio, Land-use Mix (Entropy), Walkable/Drivable/Steps Ratios, Network Betweenness & Morphological characteristics of the residential environment within a 400m buffer via OpenStreetMap. \\ 
\hline
\textbf{Accessibility} & Network Distance and Circuity to: Bus Stops, Metro Stations, Commercial, Institutional, and Recreational Blocks & Pedestrian-centric routing metrics calculated via OneMap API. \\ 
\hline
\textbf{Individual Context} & Age, Clinical Conditions (Knee OA, Fall History, Comorbid Status), LSA, FES, AFEAT , PHQ-9 & Clinical and psychometric profiles providing individual-level sensing context. \\ 
\hline
\textbf{Target Variable} & NEWS-A Score & Subjective neighborhood walkability perception (Tertile discretization: Low, Neutral, High). \\ 
\hline
\end{tabular}
\end{table}

\subsection{In-Context Learning Framework}
\subsubsection{Data Prepossessing}
Although widely used, NEWS-A perception scale lacks standardized clinical benchmarks or normative risk thresholds. To resolve this, we discretized the continuous NEWS-A sum scores into three categories, Low, Neutral, and High, using an equal-frequency binning strategy \cite{ijerph14101199}. In the absence of accepted benchmarks, we opted for the relative ranking rather absolute point values of perceptions, to enable the model to effectively differentiate between the most and least supportive features within the context of the study cohort. Additionally, this strategy addresses the challenge of class imbalance, which is essential for maintaining the stability of the classification model.

\subsubsection{In-Context Learning Workflow and SHAP-IQ Explanation}
Leveraging the ICL paradigm, this study employs TabPFN, a foundation model for tabular data, to perform the classification tasks. In this workflow, the model's weights remain frozen. To demonstrate the performance, the dataset is partitioned into an 80/20 split, where the training portion (80\%) is passed directly to the TabPFN as a Context Set. During inference, the model utilizes its self-attention mechanism to globally evaluate each query individual against this Context Set. By leveraging a pre-trained prior to capture how spatial and individual features interact, TabPFN generates class membership probabilities via Approximate Bayesian Inference. For benchmarking, traditional inductive ML models,  namely RF and XGBoost are evaluated as baselines. To ensure a rigorous comparison, both baseline models are extensively optimized using a grid-search cross-validation protocol to maximize their perdictive performance on this dataset.

To interpret the underlying logic of the model's predictions, traditional SHAP analysis is utilized which mainly captures first-order, independent feature contributions. SHAP-IQ is further leveraged to quantify the synergistic interactions between features. It is worth noting that, SHAP-IQ is uniquely compatible with TabPFN because the model's ICL paradigm renders the calculation of complex interaction values computationally inexpensive, compared to inductive ML models. This enables the simple removal of out-of-coalition features, allowing the model to be re-contextualized with the modified data points \cite{rundel2024interpretable}. By selectively omitting pairs of features, the model explicitly evaluates their joint impact on classification to uncover the underlying synergistic effects between the variables.

\section{Preliminary Results}
First, the predictive performance of the TabPFN (ICL), was evaluated against RF and XGBoost, two inductive ML models commonly utilized for interpreting the impacts of BE on human perception and behaviour. Then we discuss the results in terms of BE impacts on perceived neighborhood walkability.

\subsection{Classification Performance}
As the NEWS-A scores were discretized into three categories, the analytical task was framed as a three-class classification problem. Table \ref{tab:model_comparison} reports the performance of the tuned RF, XGBoost, and TabPFN (ICL) models. It was found that TabPFN outperformed both optimized inductive models across the evaluated metrics. While the tuned RF and XGBoost models achieved accuracies of 46.15\% and 53.85\%, respectively, TabPFN reached an accuracy of 55.77\%. In addition, TabPFN achieved a Macro F1 score of 54.89\%, which calculates the unweighted mean of the F1 scores across all classes independently. The F1 score itself represents the harmonic mean of precision and recall, providing a balanced metric that penalizes both false positives and false negatives. TabPFN's performance in Macro F1 represents a improvement over XGBoost (50.56\%) and RF (45.85\%). This comparison illustrates that TabPFN achieves this results without parameter tuning, whereas both baselines models require extensive optimization. It highlights the distinct advantage of the ICL paradigm, demonstrating its ability to mitigate small-data challenges while ensuring the high predictive fidelity required for downstream SHAP-IQ analysis.

\begin{table}[htbp]
\centering
\caption{Performance Comparison of Random Forest, XGBoost, and TabPFN (ICL)}
\label{tab:model_comparison}
\begin{tabular}{lccc}
\hline
Model & Accuracy (\%) & Auc (\%) & F1 (\%) \\ \hline
RF & 46.15 & 59.83 & 45.85 \\ \hline
XGBoost & 53.85 & 66.34 & 50.56 \\ \hline
TabPFN (ICL) & \textbf{55.77} & \textbf{76.62} & \textbf{54.89} \\ \hline
\end{tabular}
\end{table}

\subsection{Influential BE Features}
\begin{figure}[htbp]
    \centering
    \includegraphics[width=1\linewidth]{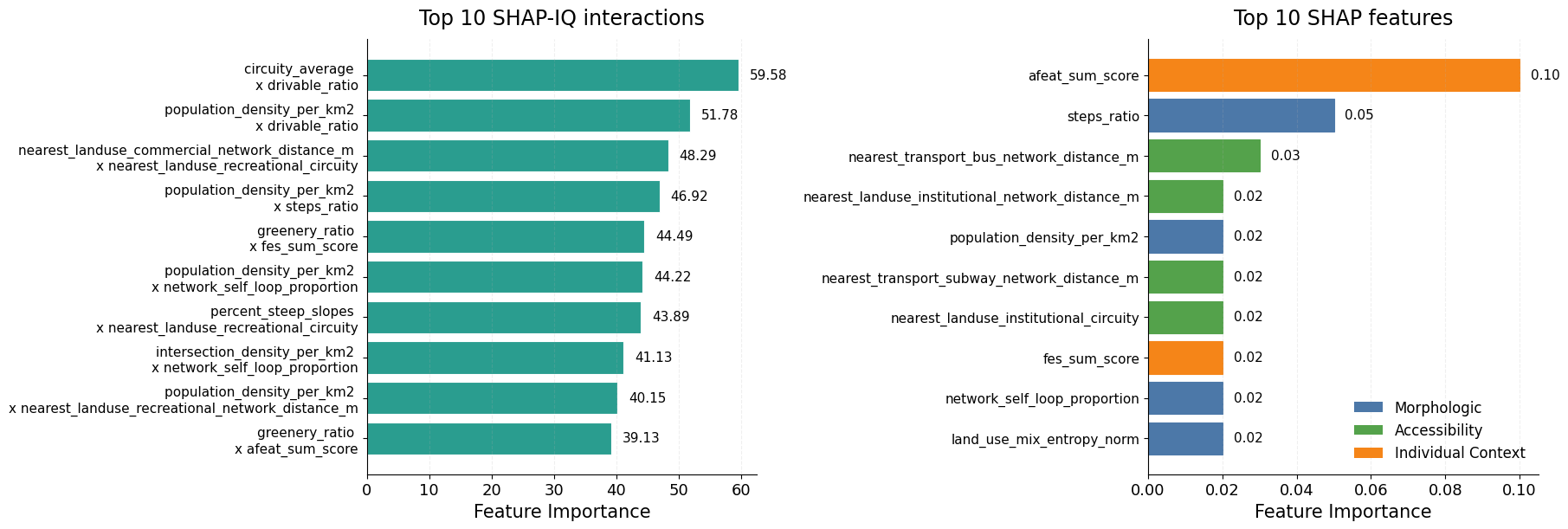}
    \caption{Top 10 feature importances for the TabPFN model. The left panel plots the SHAP-IQ synergistic effects, while the right panel displays the typical SHAP values.}
    \label{fig:shapiq}
\end{figure}

The feature importance derived from TabPFN is presented in Figure \ref{fig:shapiq}, comparing SHAP-IQ feature interactions (left) with standard isolated SHAP values (right). When evaluated in isolation, individual clinical contexts, specifically the AFEAT and FES scores, alongside localized morphological barriers like the steps ratio, emerge as the top independent predictors of perceived walkability. However, the SHAP-IQ analysis (left) reveals a critical shift in feature importance: when synergistic effects are quantified, perception is fundamentally governed by complex variable pairs rather than isolated factors. Ranking at the top of the interaction importance is the synergistic effect between the neighborhood's average walkway circuity and the ratio of drivable roads (circuity\_average $\times$ drivable\_ratio). The magnitude of this interaction reveals that the model views winding layouts in car-dominated spaces as the most critical determinant of perceived walkability. Similarly, the significant interaction importance between steep slopes and recreational circuity illustrates that the model weights topography most heavily when it is compounded by navigational complexity.  Most notably, while greenery appears unimportant in isolation, it becomes a top influential factor when interacting with FES and AFEAT scores. This highlights that greenery acts as a restorative buffer, modulating perceived walkability for mobility-impaired older adults 

\section{Discussion and Conclusion}
This study demonstrates the potential of ICL for understanding the complex interplay between the BE and the walkability perceptions of  a unique demographic: older adults with knee OA or fall histories. By leveraging TabPFN, we achieved superior classification performance in predicting neighborhood walkability perceptions compared to fine-tuned inductive ML models. The resulting feature importance analysis reveals that the model’s decisions are primarily driven by synergistic effects rather than isolated variables. The interplay between street circuity and ratio of drivable roads emerges as the primary predictive discriminator. Neighborhood greenery is found to hold substantial weight only when coupled with an individual’s fear of falling or perceptions of age-friendliness, identifying critical interactions that traditional SHAP analyses often obscure.

As an exploratory investigation into leveraging ICL for explaining the perception of neighborhood walkability, certain limitations exist. First, while SHAP-IQ identifies the hierarchical importance of feature synergies, the current analysis focuses on the magnitude of these interactions rather than their directional valence. Future work will utilize dependence analysis to explicitly characterize whether these synergies exert positive or negative impacts on perception. Second, the latent synergies identified via ICL lack cross-validation through conventional statistical methods; subsequent research should feed these extracted interactions back into standard regression models to enhance both the explainability and practical utility of the findings. Finally, the cross-sectional design and reliance on the self-reported NEWS-A scale capture perceived walkability rather than objective mobility. Future studies should supplement these subjective assessments with longitudinal GPS tracking to observe how environmental perceptions translate into actual spatial trajectories.


\bibliography{lipics-v2021-sample-article}

\appendix

\end{document}